\pdfoutput=1

\documentclass[11pt]{article}
\usepackage [ table ]{ xcolor }

\usepackage[final]{acl}

\usepackage{times}
\usepackage{latexsym}

\usepackage[T1]{fontenc}

\usepackage[utf8]{inputenc}

\usepackage{microtype}

\usepackage{inconsolata}

\usepackage{xspace}
\newcommand{\our}{CuPUL\xspace}
\usepackage{graphicx}
\usepackage{dsfont}
\usepackage{amsmath, amsfonts}
\usepackage{amsmath,amssymb}
\usepackage{pifont}
\newcommand{\thicktilde}[1]{\mathbf{\tilde{\text{$#1$}}}}
\usepackage{multirow}
\usepackage{enumitem} 
\usepackage{booktabs} 
\usepackage{tabularx} 
\usepackage{makecell} 

\title{Re-Examine Distantly Supervised NER: \\
A New Benchmark and a Simple Approach}

\author{
    Yuepei Li \quad
    Kang Zhou \quad
    Qiao Qiao \quad
    Qing Wang \quad
    Qi Li \\
    Iowa State University, Ames, Iowa, USA \\
    \texttt{\{liyp0095, kangzhou, qqiao1, qingwang, qli\}@iastate.edu} \\
  }

\begin{document}

\maketitle
\begin{abstract}
Distantly-Supervised Named Entity Recognition (DS-NER) uses knowledge bases or dictionaries for annotations, reducing manual efforts but rely on large human labeled validation set.
In this paper, we introduce a real-life DS-NER dataset, QTL, where the training data is annotated using domain dictionaries and the test data is annotated by domain experts. This dataset has a small validation set, reflecting real-life scenarios. Existing DS-NER approaches fail when applied to QTL, which motivate us to re-examine existing DS-NER approaches. We found that many of them rely on large validation sets and some used test set for tuning inappropriately. To solve this issue,
we proposed a new approach, token-level Curriculum-based Positive-Unlabeled Learning (CuPUL), which uses curriculum learning to order training samples from easy to hard. This method stabilizes training, making it robust and effective on small validation sets. CuPUL also addresses false negative issues using the Positive-Unlabeled learning paradigm, demonstrating improved performance in real-life applications.\footnote{Our code and new QTL dataset are available at \url{https://github.com/liyp0095/CuPUL}.}
\end{abstract}

\definecolor{lightgray}{gray}{0.9} 
	
\definecolor{Gray}{gray}{0.9}

\section{Introduction}\label{sec:intro}

Distantly-Supervised Named Entity Recognition (DS-NER) is a task to leverage existing knowledge bases (KBs) or dictionaries to provide annotations for named entity recognition tasks. This approach significantly reduces the need for labor-intensive manual annotations, but it faces challenges due to issues in automated annotations, such as false positives and false negatives. To address the annotation errors, various methods are proposed. Some studies focus on false negative issues \cite{shang2018learning, peng2019distantly,zhou2022distantly}.
Others propose to tackle general noisy annotations through noise removal processes \cite{meng2021distantly,liang2020bond,hedderich2018training,zhang2021denoising,liu2021noisy}. 

To assess the effectiveness of existing DS-NER approaches, 
we introduce a real-life DS-NER dataset, QTL (Quantitative Trait Locus), which is annotated for trait entities in the animal science domain. Unlike previous datasets, QTL contains a very small validation set of only 21 sentences, avoiding the significant manual effort required to obtain large validation sets in real-life scenarios. In contrast to previous benchmark datasets where entity mentions often comprise proper nouns, the trait entities in the QTL dataset are descriptive terms, such as ``tail size'' and ``hoof color''.

While existing DS-NER methods perform well on benchmark datasets such as CoNLL2003, often rivaling fully supervised approaches, they consistently fail when applied to our QTL dataset. This motivates us to re-examine existing DS-NER approaches. We identify some issues: Some approaches \cite{liang2020bond,zhang2021improving,10.1609/aaai.v37i11.26583_atsen} deviate from the DS-NER framework and directly use the test set for hyperparameter tuning, leading to unreliable performance. Some approaches \cite{shang2018learning,meng2021distantly} train their models with fixed hyperparameters, yet fail to achieve consistent results across different datasets. The remaining approaches \cite{NEURIPS2023_359ddb9c,wu-etal-2023-mproto} employ a validation set from fully supervised (FS) data for parameter tuning. These approaches overlook the significant manual labor required to obtain a validation set for parameter tuning in real-life scenarios, affecting the robustness of existing methodologies when applied to real-life applications with a small validation set, thereby compromising the reliability of these approaches.

To solve the issues mentioned above, 
we present a simple yet effective approach inspired by Curriculum learning and Positive-Unlabeled (PU) learning , named CuPUL. The motivation behind curriculum learning is that deep learning models are non-convex and trained using batches of samples, so the order of training data can significantly impact model performance. Curriculum learning rearranges the batches of training samples such that the model learns from easy to hard samples and revisits easier samples more frequently. With this new arrangement, models tend to converge to a better local optimum.
Furthermore, we design a token-level curriculum arrangement to address token-level noise in DS-NER tasks. We observe that "easy samples" are usually cleaner, and learning from these first can initially avoid label noise, making the model more robust. To tackle false negative issues, we adopt the Positive-Unlabeled learning paradigm.

In summary, our main contributions are:
\begin{itemize}
\setlength\itemsep{0em}
\item We present a real-life DS-NER dataset, QTL, and test the performance of the existing state-of-the-art methods. We observe that many methods do not follow the practical DS-NER setting and have unsatisfactory performance.

\item We propose a simple method \our to address the noise issue in DS-NER. We empirically demonstrate that \our can significantly outperform the state-of-the-art DS-NER method on the QTL dataset and different benchmark datasets. 

\end{itemize}
\section{QTL Benchmark}\label{sec:qtl}

To reduce the cost of human-annotated training data for NER tasks, DS-NER uses professional dictionaries or knowledgebases for annotations. Existing DS-NER benchmark datasets use NER benchmark datasets to simulate the distant supervision setting by replacing the human annotations on training datasets with knowledge base annotations \cite{liang2020bond, shang2018learning, zhou2022distantly}. 
Among these benchmarks, only the BC5CDR \cite{shang2018learning,li2016biocreative} dataset comes from professional domains where DS-NER tasks are in high demand.

We present \textbf{Q}uantitative
\textbf{T}rait \textbf{L}ocus (QTL), a real-life DS-NER application in the animal science domain. The entity type to recognize is ``trait'', an important task in the construction of genotype-phenotype databases for advancing livestock genomics research and breeding methodologies (examples in Table \ref{tab:traits}). 


Compared to previous DS-NER benchmark datasets, the QTL dataset presents two main distinctions: 1) The entities in QTL are generally longer in length. 2) Unlike previous benchmarks, where entities are typically proper nouns, QTL entities are often descriptive expressions. To describe the distinct characteristics of trait entities, we compare the Trait type from the QTL test set with entity types (PER, LOC, ORG, MISC) from the CoNLL03 \cite{tjong2003introduction} test set. Some key statistics are presented in Table \ref{tab:entity_statistics}, revealing that Trait entities have a longer entity length on average and a lower proper noun ratio compared to the benchmark dataset. These characteristics make the QTL dataset unique and introduce additional challenges in the DS-NER task.

\begin{table}[t]
    \centering
    \setlength{\tabcolsep}{6pt} 
    \resizebox{0.4\textwidth}{!}{
    \begin{tabular}{cl}
        \toprule
        No. & Trait Entity Example \\
        \midrule
        1 & Fatty acid composition of milk \\
        2 & Dwarf phenotype \\
        3 & Total number of piglets born per litter \\
        4 & Body mass index \\
        5 & The number of first to third births \\
        \bottomrule
    \end{tabular}
    }
    \caption{List of Trait Entities from Test Set of QTL.}
    \label{tab:traits}
\end{table}

\begin{table}[t]
    \centering
    \setlength{\tabcolsep}{6pt} 
    \resizebox{0.5\textwidth}{!}{
    \begin{tabular}{lcccc}
        \toprule
        \makecell{Dataset} & \makecell{Entity Type} & \makecell{\# Entity} & \makecell{Average Entity\\Length} & \makecell{Proper Noun\\Ratio} \\
        \midrule
                & PER & 1617 & 1.71 & 0.97 \\
        CoNLL03 & LOC & 1662 & 1.16 & 0.98 \\
                & ORG & 1656 & 1.51 & 0.92 \\
                & MISC & 693 & 1.32 & 0.60 \\
        \midrule
        QTL & Trait & 1219 & 1.98 & 0.24 \\
        \bottomrule
    \end{tabular}
    }
    \caption{Statistics on Entity Types of CoNLL2003 and QTL. The average entity length indicates the number of words per entity, and the proper noun ratio reflects the proportion of entities that contain at least one proper noun.}
    \label{tab:entity_statistics}
\end{table}

To establish the QTL dataset, we collected a corpus with 1,716 abstracts carefully selected from PubMed\footnote{\url{https://pubmed.ncbi.nlm.nih.gov/}} by domain experts for QTL studies related to six species: cattle, pig, goat, sheep, chicken, and rainbow trout. 
We randomly selected 1,609 abstracts in this corpus to establish the training data. The training data consists of 18,706 sentences with 514,176 tokens.
For the distant annotation process, the domain experts gathered a specialized dictionary of 3,884 trait names from four established domain ontologies\footnote{Vertebrate Trait (VT) Ontology, Livestock Product Trait (LPT) Ontology, Livestock Breed Ontology (LBO), and Clinical Measurement Ontology (CMO). Examples can be found at \url{https://www.animalgenome.org/QTLdb/export/trait_mappings}}. After obtaining the dictionary, string matching was used to distantly annotate the training corpus. Then from the 1609 papers, we randomly selected 21 sentences (with 952 tokens) and had a well-trained domain curator provide manual annotations to form a validation set. 

This curator later provided annotations for the remaining 107 abstracts to form a test set, which covered all six species of interest. To assess annotation quality, we had a second domain curator check the annotations on 10 randomly selected abstracts. The two curators had a total agreement. Therefore, we used the annotations as ground truth. More annotation details are described in Appendix \ref{append:data}. The test set contains 1,044 sentences with 32,251 tokens and 1,219 entities. 

Notably, the validation set is quite small in the QTL dataset. This practice followed the motivation of DS-NER tasks, where the human effort should be minimized at the training stage. This limited size of the validation set may impact the tuning of hyperparameters during the model training process, potentially affecting the model's performance. This issue reflects a realistic challenge encountered in DS-NER applications, which requires the model to be robust and not sensitive to hyperparameters.

\noindent \textbf{Annotation Limitations:} Due to the high cost, the majority of the annotations are provided by a single curator. One observation from the curators is that there is a considerable amount of discontinued trait entities. For example, in ``milk protein, lactose, and fat percentage'', there are three entities: milk protein percentage, milk lactose percentage, and milk fat percentage. Due to the annotation software limitation, this example was annotated as ``milk protein'', ``lactose'', and ``fat percentage''.

\section{Related Work Analysis}\label{sec:prelim}

\begin{table*}[t]
	\centering
	\setlength{\tabcolsep}{6pt} 
    \resizebox{\textwidth}{!}{
	\begin{tabular}{cccccccc}
        \hline
		\textbf{Method} &  \textbf{Code Provided} &  \textbf{Code Runable} & \textbf{Hyperparameter} & \textbf{Tuning required} & \textbf{Tuning Instruction}  &  \textbf{Inference model} & \textbf{Feasible}\\
		\hline
            \multicolumn{3}{l}{\textbf{DS-NER without Self-training}}\\
            \hline
            AutoNER        &  \ding{51} &  \ding{51}  & Fixed &  \ding{55} & - &  Model at Final Epoch & \ding{51} \\
            Conf-MPU        &  \ding{51} &  \ding{51} & Not Fixed &  \ding{55}  & - &  Model at Final Epoch & \ding{51} \\
            MProto        &  \ding{51} &  \ding{51}   & Not Fixed &  \ding{51}  &  \ding{55} &  Model at Final Epoch & \ding{55} \\
            \hline
            \multicolumn{3}{l}{\textbf{DS-NER with Self-training}}\\\hline
            BOND        &  \ding{51} &  \ding{51}  & Not Fixed &  \ding{51}  &  \ding{51} & Best Model on Test & \ding{51}\\
            RoSTER        &  \ding{51} &  \ding{51}  &  Fixed  & \ding{55} & - & Model at Final Epoch & \ding{51} \\
            SCDL        &  \ding{51} &  \ding{51}  & Not Fixed &  \ding{51}  &  \ding{51} & Best Model on Test & \ding{51} \\
            ATSEN        &  \ding{51} &  \ding{51}  & Not Fixed &  \ding{51}  &  \ding{51} & Best Model on Test & \ding{51} \\
            DesERT &  \ding{51} &  \ding{51} & Not Fixed  &  \ding{51}  &  \ding{51} & First Student Model & \ding{51} \\
            
            \hline
            \multicolumn{3}{l}{\textbf{Span-based DS-NER models}}\\
            \hline
            SANTA        &  \ding{51} &  \ding{51}  &  Not Fixed  &  \ding{51}  &  \ding{55} &  Model at Final Epoch & \ding{55} \\
            Top-Neg        &  \ding{51} &  \ding{55}  &  -  &  - &  - & - & \ding{55} \\
            CLIM        &  \ding{55} &  -  &  -  &  - &  - & - & \ding{55} \\
		\hline
        \end{tabular}
        }
        \caption{Feasibility Analysis of Exist Methods for DS-NER tasks.}\label{tab:feasible}
\end{table*}

\subsection{DS-NER methods}

We collected DS-NER methods published in major conferences in 2023 and their compared baselines. We categorize the existing DS-NER methods in three groups. 1) \textbf{DS-NER with Self-training}. To improve model performance, many DS-NER methods often incorporate a self-training step, utilizing mechanisms such as soft-label retraining and multi-model teacher-student frameworks. This group includes BOND \cite{liang2020bond}, RoSTER \cite{meng2021distantly}, SCDL \cite{zhang2021improving}, ATSEN \cite{10.1609/aaai.v37i11.26583_atsen} and DesERT \cite{NEURIPS2023_359ddb9c}. 2) \textbf{DS-NER without Self-training}.
This group of methods focuses on addressing the model's effectiveness in handling noise or false negatives in DS-NER. While these methods can incorporate self-training mechanisms, it is not the primary focus of these methods. This group include AutoNER \cite{shang2018learning}, Conf-MPU \cite{zhou2022distantly}, 
MProto \cite{wu-etal-2023-mproto}. 3) \textbf{Span-based DS-NER}.
The final group of methods differs from the previous two, as it is based on span-based prediction rather than sequence labeling. These methods treat each span within a sentence as the prediction target. Previous work \cite{li-etal-2023-class} has shown that span-based NER models often outperform sequence-based NER methods in terms of effectiveness, albeit at the cost of increased algorithmic complexity. This group includes Top-Neg \cite{xu-etal-2023-sampling}, CLIM \cite{li-etal-2023-class} and SANTA \cite{si-etal-2023-santa}.
More details can be found in Appendix \ref{append:baseline}.

\subsection{Method Analysis}
We first analyze the feasibility of existing DS-NER methods in real-life applications. For a method to be considered feasible, it must provide runnable code and instructions for hyperparameter tuning if necessary. Table \ref{tab:feasible} presents our feasibility analysis results base on the manuscripts and code repositories (accessed in April 2024). We find that 1) MProto and SANTA do not provide hyperparameter tuning instructions; 2) CLIM and Top-Neg do not provide runnable code; and 3) BOND, SCDL, and ATSEN selected their inference model based on performance on the test set according to their released repositories. Thus in our empirical studies, for a fair comparison, we only re-examine feasible methods and update some methods to select the inference model based on performance on the validation set only.

The motivation of DS-NER methods is that the manual annotations are too costly to obtain. Therefore, to reduce the amount of manual annotation, the annotations in the training set come from knowledge bases or dictionaries, and the manual labeled validation set should not be large either. Existing methods focus on the first setting while neglecting the importance of the second setting. We analyzed the feasible methods in Table \ref{tab:feasible} based on these DS-NER settings and have the following observations. 
First, AutoNER and RoSTER use fixed hyperparameters. These approaches do not require hyperparameter tuning, thereby avoiding the need for a validation set. Second, Conf-MPU provides a strategy for pre-selecting hyperparameters, so it does not require a validation set either. However, the remaining methods (BOND, SCDL, ATSEN, and DesSERT) need a validation set for hyperparameter tuning. The size of the validation set may affect their performance. We present a detailed analysis of this impact in Section \ref{sec:exper}.

We do not consider the few-shot NER setting in this paper. The recent SOTA performance of few-shot NER is achieved by LLMs with in-context learning (ICL), but it is unsatisfactory, particularly in specific professional domains \cite{munnangi2024fly,hu2024improving,monajatipoor2024llms}. Moreover, the high cost of LLMs makes it less feasible for large-scale annotation tasks. BERT-based DS-NER methods are easy to deploy and can achieve high performance with low cost. Therefore, we focus on the DS-NER setting using the BERT-based model.
\section{Methodology}\label{sec:methodology}
\begin{figure}[t]
  \centering
  \includegraphics[width=0.95\linewidth]{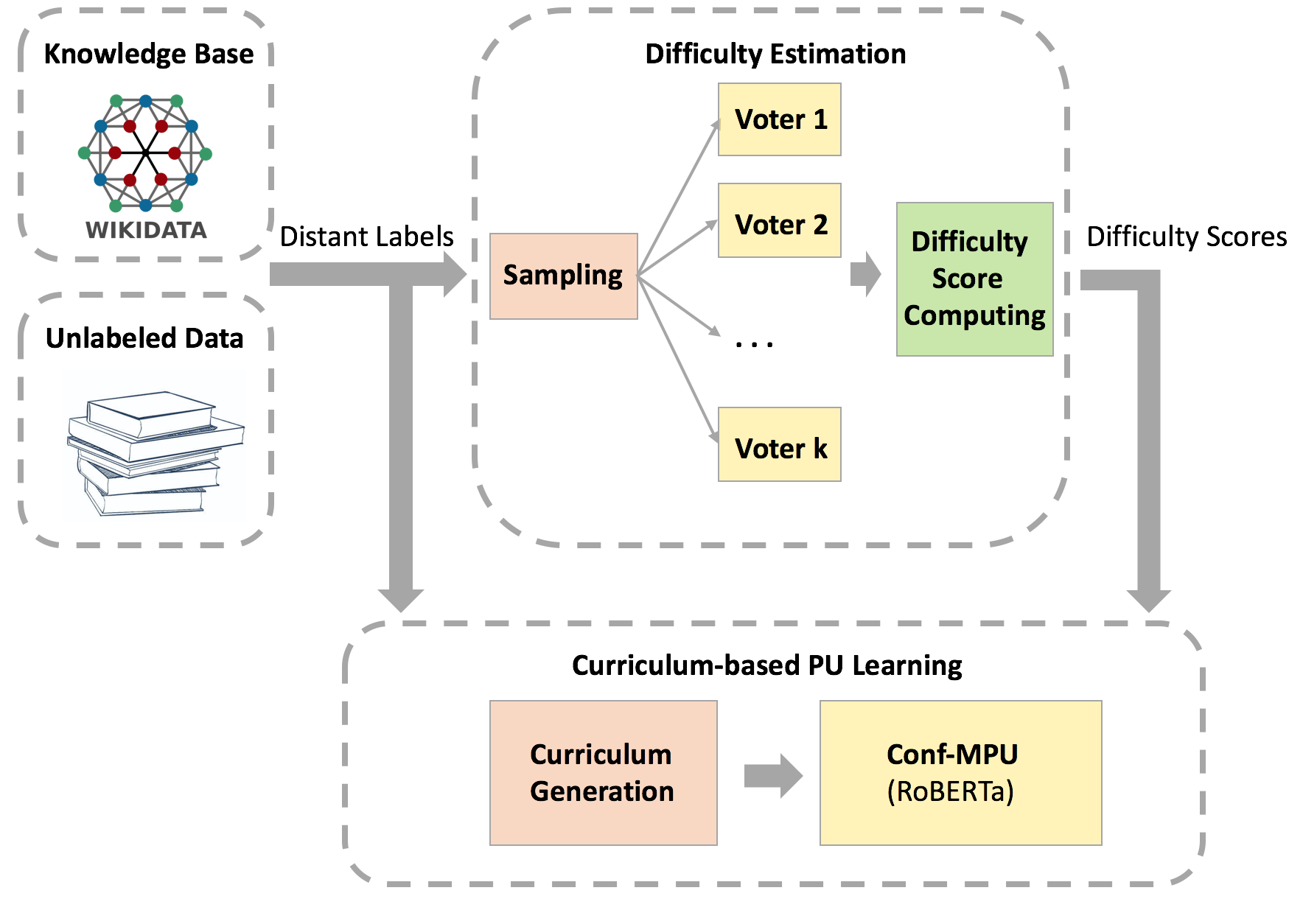}
  \caption{Overview of CuPUL}

  \label{fig: framework}
\end{figure}
In this section, we introduce a simple DS-NER method that combines the advantages of curriculum learning and PU learning. %
Figure \ref{fig: framework} shows the overview of the proposed method \our. 
The method starts by training several \textit{voters} using the distantly annotated data to calculate token difficulty scores. Then \our trains a NER classifier following the curriculum scheduler using confidence-based positive-unlabeled learning risk estimation.

\textbf{Problem Formulation:}
We denote an input sentence with $M$ tokens as $\boldsymbol{x}=[x_1, x_2, \cdots, x_M]$ and denote corresponding annotations as $\boldsymbol{y}=[y_1, y_2, \cdots, y_M]$, $y_i \in \{0, 1, \cdots, k\}$, 
where 0 denotes the unlabeled type and $1,\cdots,k$ denote $k$ entity types. For the models, a pre-trained language model such as RoBERTa is used to encode token representations and followed by a softmax function to forward the prediction of entity labels for each token in the sentence.

\subsection{Difficulty Estimation}\label{section: de}
Curriculum learning has two main steps: difficulty estimation and curriculum scheduler \cite{kocmi2017curriculum}. More details and related work of curriculum learning are discussed in Appendix \ref{sec: cl}.

Motivated by the token-level noises in DS-NER tasks, we design the difficulty estimator and the curriculum scheduler at the token level as well. It allows the model to learn from one sentence by ignoring the noisy tokens. For example, in the sentence “Peter(PER) lives(O) in(O) America(ORG)”, “Peter”, “lives”, and “in” are clean samples, and “America” is a noisy sample. The model can learn from “Peter lives in X” by ignoring the noise in the sentence. 
The token's difficulty score should reflect its inherent learnability. These scores are estimated using the disagreements between basic NER models or voters.

\subsubsection{Voters}
 
For training the voters, a neural network for NER classification is used. 
The design of the voters demands simplicity and variability. Thus, the voters are trained using a regular multi-class classification risk function. 
The training process follows the Positive-Negative setting, where 0 represents non-entity type. Label imbalance in NER tasks is mitigated by sampling negative samples. Note that the performance of the voter itself does not affect the final outcomes of CuPUL, which we will introduce in the section \ref{section:CD}.

\subsubsection{Difficulty Scores}
After training $V$ voters, each token $x$ receives $V$ predicted class probabilities $f(x, \boldsymbol{\theta}_1), ..., f(x, \boldsymbol{\theta}_V)$, where $\boldsymbol{\theta}_1 ... \boldsymbol{\theta}_V$ are the voters' parameters. The prediction $f(x, \boldsymbol{\theta}_i)$ is a vector that represents the class distribution of each token $x$ denoted as $\boldsymbol{\Pr}_i(x)$. The difficulty of the token is assessed based on the disagreement among these distributions. Specifically, we use Kullback-Leibler (KL) divergence, a measurement for dissimilarities of two distributions $\boldsymbol{\Pr}_i(x)$ and $\boldsymbol{\Pr}_j(x)$, to calculate the disagreement level of two voters. Mathematically, it is:
\begin{multline}
    H_{ij} = \frac{1}{2} \{D_{KL}(\boldsymbol{\Pr}_i(x)||\boldsymbol{\Pr}_j(x))) + \\
    D_{KL}(\boldsymbol{\Pr}_j(x))||\boldsymbol{\Pr}_i(x))\},
\end{multline}
where $D_{KL}(\cdot)$ denotes the KL divergence. KL divergence is asymmetric. By taking the average of $H_{ij}$ and $H_{ji}$, we derive a symmetric difficulty score $H_{\{ij\}}$.

Given that there are $V$ voters, the final difficulty score for each token $x$ is defined as the average of the non-identical pairs among all voters:
\begin{equation}\label{eq: difficuty score}
    H = \frac{\sum_{i=1}^{V}\sum_{j=i+1}^{V} H_{\{ij\}}}{V\cdot(V-1)/2}.
\end{equation}
Eq.(\ref{eq: difficuty score}) defines the token difficulty scores as an arithmetic mean of disagreements between pair-wise voters. Consequently, a token's difficulty score is low when all voters agree, and it increases with greater disagreement.

\subsection{Curriculum Design} \label{section:CD}



To avoid overfitting negative samples, we adopt Positive-Unlabeled (PU) learning based risk estimation, treating data labeled with 0 as unlabeled rather than non-entity. PU learning assumes the unlabeled data represents the entire dataset's distribution \cite{zhou2022distantly,zhou2023improving}. To meet this assumption, we include all unlabeled data in the first curriculum, scheduling only the labeled positive data. 

Our curriculum is based on token difficulty scores $H$, which follow a long-tail distribution, making most tokens ``easy'' (Figure \ref{fig: distribution}). Previous research \cite{platanios2019competence,GNANASHEELA2013425} indicates that a uniform difficulty range may render curriculum learning ineffective. Therefore, we propose a power-law selector for a more effective curriculum scheduler.


To build the curricula, we first arrange all $T_u$ unlabeled tokens followed by $T_p$ positive-labeled tokens sorted by their difficulty scores in ascending order. The first curriculum consists of all unlabeled tokens and the first $\tau T_p$ labeled positive tokens, where $\tau$ ($0<\tau<1$) is a selective factor. The second curriculum consists of the first $\tau^2 T$ tokens from the remaining $(1-\tau)T_p$ tokens. This selection process continues until the penultimate curriculum. The remaining tokens are placed in the final curriculum. These curricula are denoted as $C_1, C_2, ..., C_\eta$. For example, suppose $T_p=20$, $T_u=80$, $\tau=0.5$, and $\eta=3$. Then, $C_1$ consists of tokens indexed from 1 to 90 ($80$ unlabeled tokens and the $10$ easiest positive tokens), $C_2$ consists of tokens indexed from 91 to 95, and $C_3$ consists of tokens indexed from 96 to 100. 




\subsection{Curriculum-based PU Learning} \label{section:cpu}


We train the NER classifier across $\eta$ curricula using the ``Baby Step'' training schedule \cite{spitkovsky2010baby,cirik2017visualizing}. Starting with $C_1$, we add each subsequent curriculum after a fixed number of epochs, training through all curricula until completion. The training stages ($\{S_i, 1<i\leq \eta\}$) correspond to the number of curricula, with the model trained over multiple epochs in each stage. Each stage is treated as an independent training segment, with earlier curricula being reviewed more frequently, enhancing learning under PU assumptions and resulting in a robust curriculum learning framework.

Specifically, we adopt the Conf-MPU loss function, proposed by \citet{zhou2022distantly}, as the backbone PU loss function in the curriculum-based training. 

Details of Conf-MPU can be found in Appendix \ref{sec:PU}.
Instead of having entity confidence score $\lambda(x)$ estimated by another binary PU model, the only difference we make is to reuse the voters trained in Section \ref{section: de} to ensemble the confidence score for each token $x$. We use the soft-label ensemble as 
\begin{equation}
    \boldsymbol{\Pr}(x)=\frac{\sum_{j=1}^V f(x,\boldsymbol{\theta}_j)}{V},
\end{equation}
where $\boldsymbol{\Pr}(x)$ is the ensemble probability distribution over all classes. 
The confidence score of a token $x$ being an entity token is then calculated as
\begin{equation}
    \lambda(x) = \sum_{j=1}^k \boldsymbol{\Pr}_j(x).
\end{equation}

For the neural network of the NER classifier, we choose the structure described at the beginning of Section \ref{sec:methodology}.

\subsection{Self-Training}

Several studies \cite{liang2020bond, peng2019distantly, meng2021distantly} have shown that self-training can effectively upgrade the performance of a trained DS-NER model. We apply the self-training method in \citet{meng2021distantly}, which uses soft labels to conduct self-training and a masked language model to conduct contextual data augmentation simultaneously. Self-training is used directly after \our, and we call the classifier with self-training ``CuPUL+ST''.

\section{Experimental Studies}\label{sec:exper}

\subsection{Baseline Methods}


We use feasible methods mentioned in Section \ref{sec:prelim} as baseline methods. First, we report distant supervision results as KB-Matching. We classify feasible DS-NER methods into two groups. 1) \textbf{DS-NER without Self-training} consists of AutoNER \cite{shang2018learning} and Conf-MPU \cite{zhou2022distantly}. 
\our is directly comparable with these methods. We also include an ablation version of CuPUL (CuPUL-curr), which removes Curriculum Learning, as a baseline.
2) 
\textbf{DS-NER with Self-training} includes BOND \cite{liang2020bond}, RoSTER \cite{meng2021distantly}, SCDL \cite{zhang2021improving} and ATSEN \cite{10.1609/aaai.v37i11.26583_atsen} and DesERT \cite{NEURIPS2023_359ddb9c}. These methods apply teach-student or training augmentation steps to further boost the DS-NER performance. CuPUL+ST is directly comparable with these methods.

To ensure a fair comparison, we made some necessary code modifications to the baseline methods. For Conf-MPU, we updated the encoding model to RoBERTa. For BOND, SCDL, ATSEN, and DesSERT, we modified the hyperparameter tuning process to use the validation set instead of the test set. Early stopping is used to select the inference model. RoSTER uses fixed parameters, but the \texttt{max\_seq\_length} did not meet the requirements for some datasets, so we adjusted it accordingly. Specific parameters details are in Appendix \ref{append:setting}.



\subsection{QTL Experiments}
\textbf{Evaluation Metrics:} Due to the annotation limitation and the fact that none of the DS-NER methods can handle discontinued spans, we include relaxed Precision, Recall, and F1 scores to evaluate the performance on the QTL dataset, in addition to the strict span-level Precision, Recall, and F1 scores used in previous studies. For relaxed metrics, it deems a predicted span correct if there is at least one overlapping word with the ground truth annotation.  According to the curator's feedback, the relaxed metrics can meet the practical need as identifying potential entities is more important than identifying precise boundaries.



\begin{table}[t]
	\centering
	\setlength{\tabcolsep}{6pt} 
    \resizebox{0.48\textwidth}{!}{
	\begin{tabular}{ccc}
        \hline
		\textbf{Method} & \textbf{QTL-strict} &  \textbf{QTL-relax}  \\
		\hline
            \multicolumn{3}{l}{\textbf{DS-NER without Self-training}}\\
            \hline
            KB-Matching    & 37.15 (\textbf{82.95}/23.93)& 41.86 (\textbf{93.46}/26.97)\\
            AutoNER        &  41.67 (69.07/29.83) & 55.49 (83.17/41.64)\\
            Conf-MPU     & 52.07 (76.30/45.37)   & 60.58 (91.15/51.28)    \\ 
            \textbf{CuPUL-curr}  &      54.75 (75.40/42.99) & 62.94 (86.76/49.38)   \\ 
		  \textbf{CuPUL}      & \textbf{56.84} (73.03/\textbf{46.51})  & \textbf{66.18} (85.31/\textbf{54.06})   \\
          \hline
          \multicolumn{3}{l}{\textbf{DS-NER with Self-training}}\\\hline
            BOND           & 53.08 (60.89/47.04)     & 65.57 (77.97/56.57)   \\
            RoSTER          & 47.80 (73.12/35.51)      & 55.43 (\textbf{91.35}/39.79)\\
            SCDL        & 43.62 (\textbf{79.57}/30.05)& 50.18 (89.85/34.81)      \\
            ATSEN       &   46.23 (66.98/35.30)   & 51.64 (86.21/36.86)      \\
            DesERT  & 54.41 (69.20/44.83) & 64.23 (82.41/51.50)\\ 
            \textbf{CuPUL+ST}        &  \textbf{58.87} (58.28/\textbf{59.47})  & \textbf{73.57} (73.07/\textbf{74.08})   \\
		\hline
        \end{tabular}
        }
        \caption{Performance on QTL dataset: F1 Score (Precision/Recall) (in \%). The best results are in \textbf{bold}.}\label{tab:table_qtl}
\end{table}

\begin{table*}[t]
	\centering
	\setlength{\tabcolsep}{6pt} 
    \resizebox{0.88\textwidth}{!}{
	\begin{tabular}{crcccccc}
        \hline
		\textbf{Method} & & \textbf{CoNLL03} &  \textbf{Twitter} & \textbf{OntoNotes5.0} & \textbf{Wikigold} & \textbf{Webpage} & \textbf{BC5CDR} \\
		\hline
            \multicolumn{6}{l}{\textbf{DS-NER Without Self-training}} \\
		\hline
            KB-Matching & $*$ & 71.40  & 35.83  & 59.51  & 47.76 & 52.45 & 64.32 \\
            AutoNER & $*$ & 67.00  & 26.10    & 67.18   & 47.54 & 51.39  & 79.99 \\
            Conf-MPU & $\dagger$ &82.39 &43.21 &66.04 & 66.58 & 63.32 & 80.06 \\ 
            \textbf{CuPUL-curr}  &    &      83.18 & 50.12  & 67.76  & 66.43 & 65.15  & 79.29\\ 
		\textbf{CuPUL}       &     & \textbf{85.09}  & \textbf{54.34}   &  \textbf{68.06}  & \textbf{70.53}  & \textbf{73.10} & \textbf{80.19}\\
            \hline
            \multicolumn{6}{l}{\textbf{DS-NER With Self-training}} \\
		\hline
            
            RoSTER &  & ~~85.40$^*$ &~~43.91$^\dagger$ &\textbf{~~69.10}$^\dagger$ &~~58.34$^*$ &~~56.80$^\dagger$ & ~~79.78$^\dagger$\\
            \multirow{2}{*}{BOND}   
            &  $\dagger$ & 79.89    & 45.98  & 66.86 & 57.81 & 48.76 & 76.91 \\
                    & \textcolor{gray}{$*$}  & \textcolor{gray}{81.15}    & \textcolor{gray}{48.01}  & \textcolor{gray}{68.35} & \textcolor{gray}{60.07} & \textcolor{gray}{65.74} & \textcolor{gray}- \\
            \multirow{2}{*}{SCDL}   &  $\dagger$    & 82.47   &  44.76   & 68.50    & 47.62  & 41.29   & 77.72 \\
                   &  \textcolor{gray}{$*$}     & \textcolor{gray}{83.69}  & \textcolor{gray}{51.10}    & \textcolor{gray}{68.61}    & \textcolor{gray}{64.13}  & \textcolor{gray}{68.47}   & \textcolor{gray}- \\
            \multirow{2}{*}{ATSEN}  &  $\dagger$  &   79.39   & 49.38   & 68.22  & 60.72  & 43.03  & 79.95 \\
                   & \textcolor{gray}{$*$} &   \textcolor{gray}{85.59}   & \textcolor{gray}{52.46}  & \textcolor{gray}{68.95}  & \textcolor{gray}-  & \textcolor{gray}{70.55}  & \textcolor{gray}- \\
            \multirow{2}{*}{DesERT} & $\dagger$ &80.57 &48.21 &67.94 &60.32 & 62.88 & 78.21 \\
                   & \textcolor{gray}{$*$} & \textcolor{gray}{86.95} & \textcolor{gray}{52.26} & \textcolor{gray}{69.17} & \textcolor{gray}{65.99} & \textcolor{gray}{72.73} & \textcolor{gray}- \\
            \textbf{CuPUL+ST}   &    & \textbf{86.64} &  \textbf{54.78}  & 68.20    & \textbf{70.19}  & \textbf{74.48} & \textbf{80.87} \\
		\hline
        \end{tabular}
        }
        \caption{Performance on benchmark datasets with small validation sets: F1 Score (in \%). $*$ marks the results reported from the original papers, and $\dagger$ marks the results we run. The best results are in \textbf{bold}. Data in \textbf{gray} font are \textbf{NOT} used for comparative analysis as they were tuned using either the test set or an large validation set. We include these only to contrast our re-run results with previous works.}\label{tab:table_small_valid}
\end{table*}

Table \ref{tab:table_qtl} presents the results for all methods on the QTL dataset. Note that CuPUL without curriculum learning (CuPUL-curr) is essentially equivalent to Conf-MPU when there is one entity type. 
KB matching reveals that QTL annotations suffer from low recall but have relatively high precision. We observe that DS-NER baselines without self-training have limited recall improvement, resulting in weak performance. DS-NER baselines with self-training improve recall compared to AutoNER, but still generally under-perform compared to PU-based methods. CuPUL+ST can further boost the recall compared to CuPUL, significantly outperforming all baseline methods. Specifically, strict F1 and relaxed F1 of CuPUL+ST outperform the runner-up by 5.79\% and 8.00\%, respectively. 


\subsection{Benchmark Experiments}

We also re-examine all methods on existing benchmark datasets.
\subsubsection{Datasets and Metrics}

\begin{table}
    \centering
    \resizebox{0.48\textwidth}{!}{%
    \begin{tabular}{c|cccc|c}
    \hline
         \textbf{Dataset} & & \textbf{Train}  &\textbf{Valid}& \textbf{Test} & \textbf{Types}\\
         \hline
         \multirow{2}{*}{CoNLL03}
         & Sentence &14041  &20& 3453  & \multirow{2}{*}{4} \\
         & Token & 203621    &475& 46435 & \\
         \hline
         \multirow{2}{*}{Twitter}
         & Sentence & 2393   &50& 3844 & \multirow{2}{*}{10}\\
         & Token & 44076   &719& 58064 & \\
         \hline
         \multirow{2}{*}{OntoNotes5.0}
         & Sentence &  115812   &50& 12217  & \multirow{2}{*}{18}\\
         & Token & 2200865   &1090& 230118 & \\
         \hline
         \multirow{2}{*}{Wikigold}
         & Sentence &  1142    &20& 274  & \multirow{2}{*}{4}\\
         & Token & 25819   &579& 6538 & \\
         \hline
         \multirow{2}{*}{Webpage}
         & Sentence & 385   &20& 135 & \multirow{2}{*}{4}\\
         & Token & 5293  &120& 1131 & \\
         \hline
         \multirow{2}{*}{BC5CDR}& Sentence& 4560& 20& 4797&\multirow{2}{*}{2}\\
         & Token& 118170& 533& 124750&\\
         \hline
         \multirow{2}{*}{QTL}& Sentence& 18706& 21& 1044&\multirow{2}{*}{1}\\
         & Token& 514176& 952& 32251&\\
         \hline
    \end{tabular}
    }
    \caption{The statistics of involved DS-NER datasets, the valid set comprises a small subset from the original dataset, whereas the train set and test set utilize the entire original dataset. }\label{tab:table_statistics}
\end{table}

\textbf{Datasets:} We conduct experiments on six existing benchmark datasets including CoNLL03 \cite{liang2020bond}, Twitter \cite{liang2020bond}, OntoNotes5.0 \cite{liang2020bond}, Wikigold \cite{liang2020bond}, Webpage \cite{liang2020bond}, and BC5CDR \cite{shang2018learning}.
The first five are open-domain datasets, and BC5CDR is the bio-medical domain. More details and the statistics of these datasets are summarized in Appendix \ref{append:data}. The statistics of the baseline datasets are shown in Table \ref{tab:table_statistics}. 

\noindent
\textbf{Metrics:} We use span-level Precision (P), Recall (R), and F1 scores as the evaluation metrics for all the datasets. These metrics require exact matches between predicted and actual entities. A continuous span with the same label is considered a single entity during inference.  

\noindent
\textbf{Settings:} For the benchmark dataset, we use small subsets of the validation set to tune the hyperparameters including learning rate, epochs, etc, to simulate the real-life DS-NER application scenarios. Detailed settings and statistics of the validation set can be found in Appendix \ref{append:setting}. 


\subsubsection{Results on Benchmark Datasets}


Table \ref{tab:table_small_valid} presents the overall span-level F1 scores for all feasible and proposed methods on benchmark datasets. Note that RoSTER was tested on a different version of the OntoNotes5.0 dataset \cite{meng2021distantly}. Therefore, we re-run the code on OntoNotes5.0, too. We also add the results reported from previous papers for methods BOND, SCDL, ATSEN, and DesERT as a reference to the re-run results.
We have the following observations.


\textbf{DS-NER Without Self-training.} 
From Table \ref{tab:table_small_valid}, it is obvious that KB-Matching generally exhibits low recall and, on four of the benchmark datasets, low precision as well. In contrast, noise-aware DS-NER models significantly outperform KB-Matching. Furthermore, CuPUL achieves the best F1 scores on all datasets compared to all DS-NER models without self-training. The results of CuPUL-curr are very similar to those of Conf-MPU, except for the Twitter dataset. This difference is due to CuPUL using a different loss function to train the model that obtains the confidence score for each token. For NER tasks with more than 10 entity types (Twitter and OntoNotes5.0), we opted for cross-entropy instead of MAE as the loss function, which has proven to be effective. A detailed discussion can be found in Appendix \ref{appendix: loss}.

\textbf{DS-NER With Self-training.}
The results for CuPUL+ST shown in Table \ref{tab:table_small_valid} further indicate that adding a self-training phase can enhance the performance of the CuPUL model in general. When compared with baseline DS-NER models that incorporate self-training, CuPUL+ST demonstrates superior performance on five out of six datasets. 
On the OntoNotes5.0 dataset, almost all noise-aware DS-NER models have similar performances, implying that distant annotations may contain biases difficult for the models to address.

When comparing the results of BOND, SCDL, ATSEN, and DesSERT from their original papers with our re-run results, we can observe a significant decline, especially on Twitter, Wikigold, and Webpage datasets. Because these datasets are relatively small, using small validation sets may lead to more instability in the training process and higher difficulty in selecting an appropriate inference model. The results indicate that these methods may not be robust in real-life applications. However, curriculum learning, which progresses from ``easy'' to ``hard'' samples, could stabilize the training process, making it more robust and less parameter-sensitive.

\begin{figure*}[t]
  \centering
  \includegraphics[width=0.99\linewidth]{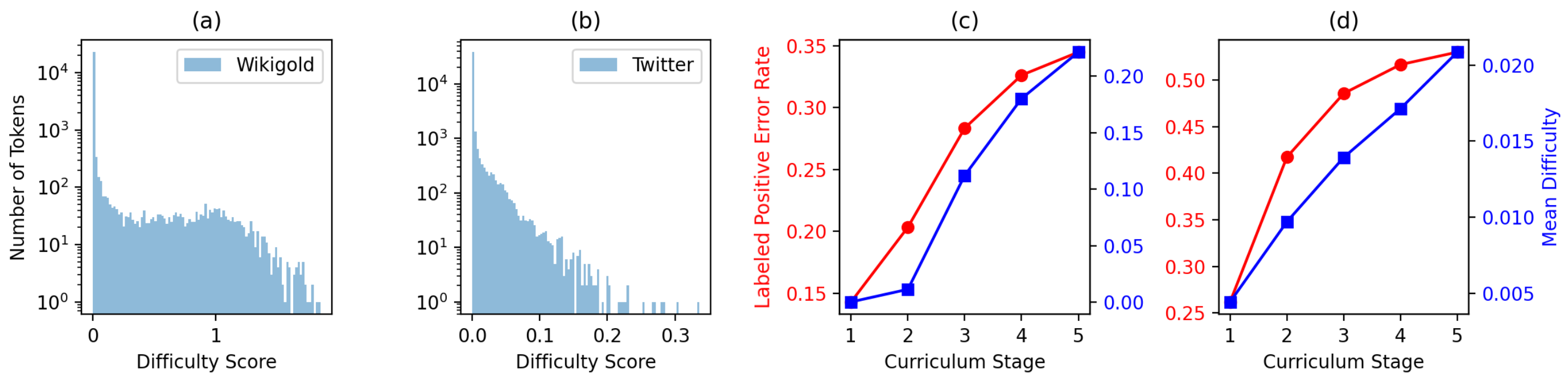}
  \caption{CuPUL Analysis: (a)(b) are the Difficulty Scores Distribution of Wikigold and Twitter, (c)(d) are the Token Level Positive Error Rate and Mean Difficulty Scores for Each Curriculum Stage on Wikigold and Twitter. }
  \label{fig: distribution}
\end{figure*}

\begin{table*}[t]
	\centering
	\setlength{\tabcolsep}{6pt} 
    \resizebox{0.9\textwidth}{!}{
	\begin{tabular}{crcccccc}
        \hline
		 & & \textbf{CoNLL03} &  \textbf{Twitter} & \textbf{OntoNotes5.0} & \textbf{Wikigold} & \textbf{Webpage} & \textbf{BC5CDR} \\
   \hline
		\textbf{CuPUL on Valid1}       &     & 85.09  & 54.34  &  68.06  & 70.53  & 73.10 & 80.19\\
		\textbf{CuPUL on Valid2}       &     & 84.55  & 54.13  &  68.25  & 68.69  & 71.48 & 80.84\\
		\hline
        \end{tabular}
        }
        \caption{Performance on benchmark datasets with different small validation sets: F1 Score (in \%). }\label{tab:table_small_valid2}
\end{table*}

\subsection{Further Analysis}

\subsubsection{Robustness of CuPUL}

To validate the robustness of CuPUL facing a small validation set, we re-selected small validation sets with the same number of sentences to train CuPUL again across CoNLL03, Twitter, Ontonotes5.0, Wikigold, Webpage, and BC5CDR datasets. 
We named the new validation sets as valid2 and the original sets as valid1, and Table \ref{tab:table_small_valid2} presents the results. The results show a slight decrease in performance on the CoNLL03, Twitter, Webpage, and Wikigold datasets and a slight increase on Ontonotes5.0 and BC5CDR datasets. Despite these fluctuations, CuPUL still outperforms all the DS-NER Without Self-training baseline methods on all datasets and DS-NER With Self-training baseline methods on most datasets compared to results in Table \ref{tab:table_small_valid}, confirming the robustness of CuPUL with small validation sets.

\subsubsection{Effectiveness}
To further validate the effectiveness of CuPUL, we conduct additional analyses using benchmark datasets. 

One important assumption we adopt for the design of curricula is that the difficulty scores follow a long-tail distribution. We illustrate the distribution of difficulty scores estimated on the Wikigold and Twitter datasets in Figure \ref{fig: distribution} (a)(b). It clearly demonstrates the long-tail phenomenon, with most tokens having low difficulty scores. 

Another important assumption adopted in CuPUL is that difficulty scores can reflect the quality of distant supervision, where ``easier'' tokens have ``cleaner'' labels. To validate this assumption and evaluate the quality of the difficulty score estimation, we examine the correlation between the difficulty scores and the quality of distant labels. We use Wikigold and Twitter as the testbed, and the results are illustrated in Figure \ref{fig: distribution} (c)(d).

For each training curriculum, we compute the token-level positive error rate (positive errors include false positives and positive type errors), and plot the rate using the left y-axis. We also compute the average difficulty scores for tokens in each curriculum shown with the right y-axis. It is clear to see that both the average token difficulty scores and positive error rate have a clear increase with respect to the order of curricula. The figure also illustrates a strong correlation between between error rate (red) and mean difficulty (blue). 
This indicates that “easy” data (data from earlier stages) have lower error rates in distant labels, meaning they are cleaner, while “hard” data (data from later stages) have higher error rates in distant labels, indicating they are noisier.
The clean data can initialize the model with a better starting point and improve the model's robustness to noise in the latter curricula.

\subsubsection{Ablation Study}

To evaluate the effectiveness of curriculum learning in CuPUL, we compare it with two variations shown in Table \ref{tab:table_ablation}. First, we use a voter ensemble, averaging predictions from five voters trained on positive and sampled negative examples. Second, we include \our-curr, which is \our without curriculum learning described in Table \ref{tab:table_small_valid}, as another variation. To assess the impact of Conf-MPU loss estimation, we compare it with a regular loss estimation (denoted as w/o Conf-MPU in Table \ref{tab:table_ablation}), where unlabeled tokens are treated as non-entity tokens. Our analysis shows that removing any component significantly lowers the F1 score. CuPUL-curr consistently achieves higher recall than w/o Conf-MPU, due to Conf-MPU addressing false negatives and partial false positives, which results in more tokens being predicted as entities. Conversely, w/o Conf-MPU achieves higher precision by addressing both false positives and positive type errors, with a more notable improvement in precision. We also observed an interesting synergistic effect on the Wikigold dataset: CuPUL achieves significantly higher precision than both w/o Conf-MPU and CuPUL-curr, as shown in Table \ref{tab:table_ablation}. The Appendix \ref{append:ablation} provides a detailed explanation and more ablation studies. The Parameter Study is discussed in Appendix \ref{section: para_study}.

\begin{table}
\setlength{\tabcolsep}{6pt}
    \centering
    \resizebox{0.48\textwidth}{!}{
    \begin{tabular}{ccccccc}
    \hline
        \multirow{2}{*}{\textbf{Method}} &  \multicolumn{3}{c}{Wikigold} &  \multicolumn{3}{c}{Twitter}\\
        \cline{2-4}  \cline{5-7}
        & \textbf{Precision} & \textbf{Recall} & \textbf{F1} & \textbf{Precision} & \textbf{Recall} & \textbf{F1} \\
         \hline
         \hline
         CuPUL &  67.06 &   74.39  &  70.53 &  54.47 &  54.20  &  54.34 \\
         \hline
         \multicolumn{4}{l}{w/o {\footnotesize Curriculum Learning}}\\
         \hline
         {\footnotesize voter ensemble} &  56.88  &  74.88  &  64.65 &  35.52  &  49.52  &  41.37 \\
        {\footnotesize CuPUL-curr} &  58.89 &   76.18   & 66.43 &  47.48 &   53.07   & 50.12 \\
        \hline 
         w/o {\footnotesize Conf-MPU} &  59.31  &  75.86  &  66.57 &  58.91  &  47.04  &  52.53\\

         \hline
    \end{tabular}
    }
    \caption{\small{Ablation study on Wikigold and Twitter datasets. \our is compared with variations without Curriculum Learning (voter ensemble only and Conf-MPU only) and  without Conf-MPU loss in Curriculum Learning.}}\label{tab:table_ablation}
\end{table}

\section{Conclusion and Future Work}\label{sec:conclusion}

In this paper, we introduce a real-life DS-NER dataset, named QTL, from the animal science domain application. We reveal the limitations of current DS-NER methods in practical DS-NER settings on the QTL dataset. To solve this issue, we propose a simple yet effective token-level curriculum-based PU learning (\our) method, which strategically orders the training data from easy to hard. Our experiments show that \our not only mitigates the adverse effects of noisy labels but also achieves state-of-the-art DS-NER on many datasets. Through \our, we demonstrate the effectiveness of curriculum learning in improving the performance of DS-NER systems. 

\clearpage
\section*{Limitations}

The limitations of the new benchmark dataset, QTL, are discussed in Section \ref{sec:qtl}.

 The "Baby Step" strategy in curriculum learning involves multiple repetitions of the first curriculum. Coupled with our power-law selector and curriculum scheduler, which tends to choose a larger initial curriculum, this may negatively impact efficiency if many curricula are established since the larger curriculum is repeatedly trained.




\section*{Ethics Statement}
We comply with the ACL Code of Ethics.

\section*{Acknowledgements}

The work is supported by NIFA grant no. 2022-67015-36217 from the USDA National Institute of Food and Agriculture and the US National Science Foundation under grant NSF-CAREER 2237831. The authors sincerely thank Cari Park for her valuable contribution to the annotation of the data.

\clearpage
\bibliography{latex/acl_latex}

\appendix
\clearpage
\section*{Appendix}

Within this supplementary material, we elaborate on the following aspects:

\begin{enumerate}
    \item \textbf{Additional Methodological Details}
    \begin{itemize}
        \item \ref{sec: cl} Curriculum Learning
        \item \ref{sec:PU} Conf-MPU Risk Estimation
        \item \ref{appendix: loss} Discussion of Loss Function
    \end{itemize}
    \item \textbf{Additional Experimental Details}, and
    \begin{itemize}
        \item \ref{append:baseline} Baselines
        \item \ref{append:data} Datasets
        \item \ref{append:setting} Hyperparameters and Experiment Settings
        \item \ref{sec:re-examine} Re-Examine Baselines on QTL
    \end{itemize}
    \item \textbf{Supplementary Experiments}
    \begin{itemize}
        \item \ref{append:ablation} Ablation Study
        \item \ref{section: para_study} Parameter Study
        \item \ref{appendix:effi} Efficiency Analysis
    \end{itemize}
\end{enumerate}

\section{Curriculum Learning}\label{sec: cl}
Curriculum learning was first proposed by \citet{bengio2009curriculum} under the assumption that learning with reordering from ``easy'' samples to ``hard'' samples would boost performance. It has been applied in various applications, including neural machine translation \cite{zhou2020uncertainty, platanios2019competence,zhou2020uncertainty,wang2018denoising}, relation extraction \cite{huang-du-2019-self,wang2023improving}, reading comprehension \cite{tay-etal-2019-simple}, natural language understanding \cite{xu-etal-2020-curriculum} and named entity recognition \cite{jafarpour-etal-2021-active, lobov-etal-2022-applying, wenjing-etal-2021-improving}.



Several studies aim to adopt curriculum learning philosophy for textual data and propose various difficulty-scoring functions and curriculum schedulers. Some methods measure sample difficulty with features derived from lexical statistics, e.g., sentence length and word rarity \cite{platanios2019competence,jafarpour-etal-2021-active}, where longer sentences and rarer words are considered ``hard''. Others use features from pre-trained language models \cite{zhou2020uncertainty,wang2018denoising,liu2020norm}. Most schedulers select samples with difficulty scores lower than a threshold  \cite{platanios2019competence}. While \citet{zhou2020uncertainty} design a sample selecting function based on model uncertainty. 
Our approach, unique in applying token-level curriculum learning to DS-NER tasks, diverges from common sentence-level methods by utilizing Transformer-based models like BERT for context-aware token-specific predictions and gradient learning.

\section{Conf-MPU Risk Estimation} \label{sec:PU}
Conf-MPU loss function has been shown to be more robust to PU assumption violation in practice. Conf-MPU estimates the risk as 
\vspace{-2mm}
\begingroup\makeatletter\def\f@size{9}\check@mathfonts
\def\maketag@@@#1{\hbox{\m@th\large\normalfont#1}}%
\begin{equation} \label{eq: conf-mpu}
        \mathrm{R}(f) = \sum_{i=1}^{k}\pi_i \big(\mathrm{R}_{{\rm{P}}_i}^{+}(f) + \mathrm{R}_{\thicktilde{\rm{P}}_i}^{-}(f) - \mathrm{R}_{{\rm{P}}_i}^{-}(f)\big) + \mathrm{R}_{\thicktilde{\rm{U}}}^{-}(f),
\end{equation}
For stage $S^*$, the number of token selected for class $i$ is $T_{i}^{S^*}$. For simplification, we denote it as $T_i^*$. The empirical estimator of Eq.(\ref{eq: conf-mpu}) is 
\vspace{-1mm}
\begingroup\makeatletter\def\f@size{9}\check@mathfonts
\def\maketag@@@#1{\hbox{\m@th\large\normalfont#1}}%
\begin{align} \label{eq8}
    &\hat{\mathrm{R}}_{\rm Conf-MPU}(f) = \sum_{i = 1}^{k} \frac{\pi_i}{T_i^*} \sum_{j=1}^{T_i^*} {\rm max}\bigg\{0, \ell (f(x_{j}^{T_i^*}, \boldsymbol{\theta}), i) \nonumber \\
    & \quad + \mathds{1}_{\hat{\lambda}(x_{j}^{T_i^*}) > \epsilon} \ell (f(x_{j}^{T_i^*}, \boldsymbol{\theta}), 0)\frac{1}{\hat{\lambda}(x_{j}^{T_i^*})} - \ell (f(x_{j}^{T_i^*}, \boldsymbol{\theta}), 0) \bigg\} \nonumber\\
    & \quad + \frac{1}{T_0^*} \sum_{j = 1}^{T_0^*} \left[\mathds{1}_{\hat{\lambda}(x_{j}^{T_0^*}) \leq \epsilon} \ell (f(x_{j}^{T_0^*}, \boldsymbol{\theta}), 0)\right], 
\end{align}\endgroup
with a non-negative constraint inspired by \citet{kiryo2017positive} ensuring the risk on the negative class. We follow \citet{zhou2022distantly} and set $\epsilon$ to 0.5 by default. 

\section{Discussion of Loss Function} \label{appendix: loss}

Two loss functions are popularly used for the DS-NER tasks. The first loss function is cross entropy (CE) loss:
\begin{equation}
    \mathcal{\ell}_{CE} = \textit{log} \: f_{i,y_i}(x;\boldsymbol{\theta}),
\end{equation}
where $f_{i,y_i}(x;\boldsymbol{\theta})$ is the prediction of token $x_i$ on class $j$.

Another commonly used loss function is mean absolute error (MAE):
\begin{equation}
    \mathcal{\ell}_{MAE} = |\boldsymbol{y}_i-f_{i,y_i}(x;\boldsymbol{\theta})|, 
\end{equation}
where $|\cdot|$ is L-1 norm of the vector and $\boldsymbol{y_i}$ denotes the one hot vector of $y_i$. 

Comparing the two loss functions, $\mathcal{\ell}_{CE}$ is unbounded, and it grants better model convergence when trained with clean data (\textit{i.e.,} $y$ are ground truth labels) because more emphasis is put on difficult tokens. However, when the labels are noisy, training with the cross-entropy loss can cause overfitting to the wrongly labeled tokens.
$\mathcal{\ell}_{MAE}$ is more noise-robust than $\mathcal{\ell}_{CE}$. It is bounded and treats every token more equally for gradient update, allowing the learning process to be dominated by the correct majority in distant labels. However, using $\mathcal{\ell}_{MAE}$ for training deep neural models generally worsens the convergence efficiency and effectiveness due to the inability to adjust for challenging training samples. 

Considering the different characteristics of these two loss functions, in practice, we suggest using $\mathcal{\ell}_{CE}$ loss for tasks with more entity types and using $\mathcal{\ell}_{MAE}$ loss for tasks with fewer number of entity types.

\section{Baselines} \label{append:baseline}
Here, we give a short description of all the baseline methods:
   \textbf{KB-Matching} distantly labels the test sets using distant supervision, serving as a reference to illustrate the performance improvements given by other advanced DS-NER methods.
    
    \textbf{AutoNER} \cite{shang2018learning} trains the neural model with a ``Tie or Break'' tagging scheme for entity boundary detection and then predicts entity type for each candidate. 
    
    \textbf{Conf-MPU} \cite{zhou2022distantly} treats the NER task as a Positive-Unlabeled learning problem and utilizes the pre-learned confidence scores to enhance the model's performance.

    \textbf{CLIM}  \cite{li-etal-2023-class} addresses the imbalance problem in the high-performance and low-performance classes by improving the candidate selection and label generation. 

    \textbf{SANTA} \cite{si-etal-2023-santa} dealing with inaccurate and incomplete annotation noise in DS-NER by utilizing separate strategies. 

    \textbf{Top-Neg} \cite{xu-etal-2023-sampling} selectively uses negative samples with high similarity to positives of the same entity type, improving performance by effectively distinguishing false negatives.
    
    
    \textbf{BOND} \cite{liang2020bond} trains a RoBERTa model on distantly labeled data with early stopping and then uses a teacher-student framework to iteratively self-train the model. 
    
    \textbf{RoSTER} \cite{meng2021distantly} employs a noise-robust loss function and a self-training process with contextual augmentation to train a NER model.
    
    \textbf{SCDL} \cite{zhang2021improving} conducts self-collaborative denoising with teacher-student framework. It trains two teacher-student networks, and the final reports come from the best model (teacher or student).

    \textbf{ATSEN} \cite{10.1609/aaai.v37i11.26583_atsen} develops a teacher-student framework with adaptive teacher learning and fine-grained student ensembling. 

    \textbf{MProto} \cite{wu-etal-2023-mproto} 
    represents each entity type with multiple prototypes to characterize the intra-class variance among entity representations and propose a noise-robust prototype network.

    \textbf{DesERT} \cite{NEURIPS2023_359ddb9c} 
    propose a novel self-training framework which augments the NER predicative pathway to solve innate distributional-bias in DS-NER.

\section{Datasets} \label{append:data}
To annotate the QTL dataset, domain experts use an online tool named TeamTat\footnote{\url{https://www.teamtat.org/}}. The screenshot of the tool is shown in Figure \ref{fig: ex_annotation}.

Here, we give a short description of the six benchmark datasets as follows:
\begin{itemize}
\setlength\itemsep{0em}
    \item CoNLL03 \cite{tjong2003introduction} is built from 1393 English news articles and consists of four entity types: person, location, organization, and miscellaneous. 
    \item Twitter \cite{godin2015multimedia} is from the WNUT 2016 NER shared task and consists of 10 entity types.
    \item OntoNotes5.0 \cite{weischedel2013ontonotes} is built from documents of multiple domains like broadcast conversations, web data, etc. It consists of 18 entity types. 
    \item Wikigold \cite{balasuriya2009named} is built from a set of Wikipedia articles (40k tokens). They are randomly selected from a 2008 English dump and manually annotated with four entity types same as CoNLL03.
    \item Webpage \cite{ratinov2009design} comprises personal, academic, and computer science conference web pages. It consists of 20 web pages that cover 783 entities with four entity types same as CoNLL03 too.
    \item BC5CDR \cite{li2016biocreative} comes from the biomedical domain. It consists of 1,500 articles, containing 15,935 Chemical and 12,852 Disease mentions.
\end{itemize}

Distant labels for benchmark datasets come from previous papers \citet{liang2020bond} and \citet{shang2018learning}.

For CoNLL03, Twitter, OntoNotes5.0, Wikigold and Webpage, the distant labels for training come from \citet{liang2020bond}. Specifically, the authors first identify potential entities through POS tagging using tools such as NLTK. They then match these potential entities by using the Wikidata query service. They use SPARQL to query the parent categories of an entity in the knowledge tree. They continue querying to the upper levels until a category corresponding to a type is found. For ambiguous entities (e.g., those associated with multiple parent categories), they are discarded during the matching process, and they are assigned type O.

For BC5CDR, the distant labels come from \citet{shang2018learning}. Specifically, the authors of the paper collect a dictionary for the biomedical dataset. The dictionary is a combination of both the MeSH database and the CTD Chemical and Disease vocabularies. The dictionary contains 322,882 chemical and disease entity surfaces. Entity labels are generated by exact string matching, and any conflicting matches are resolved by maximizing the total number of matched tokens.

\begin{figure*}[t]
  \centering
  \includegraphics[width=1.0\linewidth]{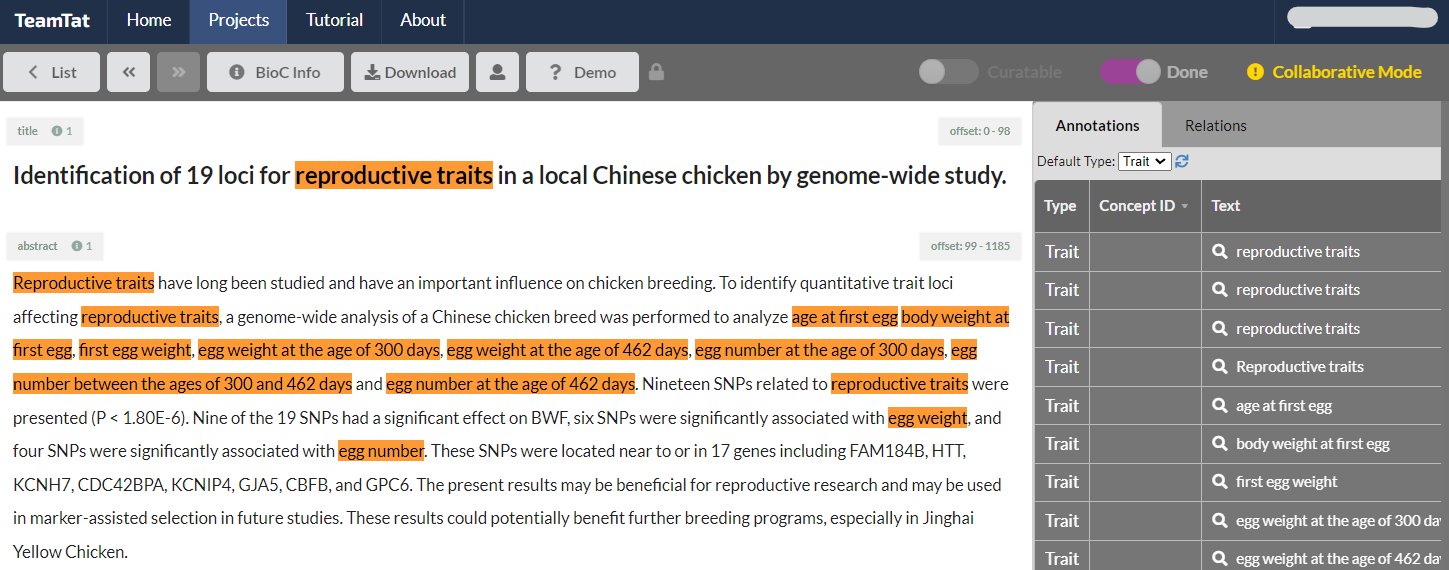}
  \caption{Screenshot for online annotation tool TeamTat.}
  \label{fig: ex_annotation}
\end{figure*}

\section{Hyperparameters and Experiment Settings} \label{append:setting}
Detailed hyper-parameter settings for each dataset are shown in Table \ref{tab:table_para}. We tune hyperparameters with Grid-Search over the small validation sets shown in Table \ref{tab:table_statistics}. Specifically, we first tune voter hyperparameters with one voter. The learning rates are set as 1e-5 for all datasets. Voter drop negative ratios are chosen from \{0.1, 0.3, 0.5\}, voter training epochs from \{1, 5, 10, 15\}, $\gamma$ from \{10, 20\}. Then we tune curriculum learning hyperparameters. The stage epochs are chosen from \{1, 2, 3\} and learning rates are chosen from \{1e-5, 3e-5, 5e-5, 7e-5, 9e-5\}. Other hyperparameters are set without tuning accordingly. For example, 
for datasets CoNLL03, OntoNotes5.0, Webpage, Twitter, Wikigold, QTL and BC5CDR, the maximum sequence length is set as 150, 230, 120, 160, 120, 180, 280 respectively, to ensure the algorithm works correctly. For all the datasets, we train them with a batch size of 32 sentences and apply Adam optimizer \cite{kingma2015adam}. The number of voters $K$ and the number of curricula $C$ are set as 5 and 5, respectively. The curriculum selective factor $\tau$ is set to 0.5 and random seed to 42. We apply cross-entropy loss to OntoNotes5.0 and Twitter since they have more entity types and apply MAE loss to other datasets.

We use the pre-trained RoBERTa as the backbone model for both the Voter and NER classifier\footnote{We will release code upon paper acceptance.}.  For all datasets, we use \textit{roberta-base}\footnote{\url{https://huggingface.co/roberta-base}}.
We report single-run results for the model performance and the random seed is set to 42. We employ PyTorch\footnote{\url{https://pytorch.org/}} and conduct all experiments on a server with a Tesla A100 GPU (32G).

\begin{table*}[t]
    \centering

    \resizebox{\textwidth}{!}{
	\begin{tabular}{c|ccccccc}
        
        \hline
		\textbf{hyper-parameter} & \textbf{CoNLL03} &  \textbf{Twitter} & \textbf{OntoNotes5.0} & \textbf{Wikigold} & \textbf{Webpage} & \textbf{BC5CDR} & \textbf{QTL} \\
        \hline
        \hline
        train set sentence \# & 14041 & 2393 & 115812 & 1142 & 385& 4560& 18706\\
        \hline
        \hline
        voter drop negative & 0.3 & 0.1 & 0.3 & 0.1&0.1&0.3&0.3 \\
        \hline
        voter learning rate &1e-5&1e-5&1e-5&1e-5&1e-5&1e-5&1e-5 \\
        \hline
        voter learning epochs &1&5&1&10&15&5&1 \\
        \hline
        Conf-MPU $\gamma$ &20&10&20&10&10&20&20 \\
        \hline
        curriculum learning stage epochs &1&2&1&2&2&1&1 \\
        \hline
        curriculum learning learning rate &1e-5&7e-5&3e-5&1e-5&5e-5&1e-5&5e-5 \\
        \hline
        
    \end{tabular}
    }
    \caption{The hyper-parameters used in CuPUL}\label{tab:table_para}
\end{table*}

\section{Re-Examine Baselines on QTL}\label{sec:re-examine}
We have explored various DS-NRE methods for QTL dataset. Our first attempt is AutoNER, which requires not only a dictionary for entity annotation but also a larger dictionary, called full-dict, for marking unknown labels, which leads to increased manual effort. To address this, we gathered a comprehensive dictionary of 26,620 potential trait entities. Unlike traditional machine learning approaches, AutoNER uses both a validation set and a test set during training and eliminates the need for hyperparameter tuning. 
In our exploration of RoBERTa-ES and BOND, we encountered the practice of using the test set for hyperparameter tuning during training. To rectify this, we modified the code to perform hyperparameter tuning on the validation set and conducted tests on the test set, focusing on hyperparameter tuning of early stop criteria and self-training period. 
For SCDL and ASTEN, we applied the hyperparameter tuning strategies outlined in the paper. Note that CuPUL without curriculum learning is essentially equivalent to Conf-MPU when there is one entity type. Therefore, Conf-MPU is not presented in the results. 




\begin{table*}[t]
	\centering
	\setlength{\tabcolsep}{6pt} 

    \resizebox{\textwidth}{!}{
	\begin{tabular}{ccccccc}
        \hline
		\textbf{Method} & \textbf{CoNLL03} &  \textbf{Twitter} & \textbf{OntoNotes5.0} & \textbf{Wikigold} & \textbf{Webpage} & \textbf{BC5CDR} \\
		\hline
            \multicolumn{6}{l}{\textbf{Fully Supervised}} \\
		\hline
            RoBERTa$^\#$           &  90.11 (89.14/91.10)   & 52.19 (51.76/52.63)   & 86.20 (84.59/87.88)   & 86.43 (85.33/87.66)   & 72.39 (66.29/79.73) &  90.99 (-/-)$^\dagger$   \\
            \hline
            \multicolumn{6}{l}{\textbf{Span-based DS-NER models}} \\
            \hline
            SANTA$^\Diamond$  &  86.59 (86.25/86.95) & - & 69.72 (69.24/70.21) & - & 71.79 (78.40/66.72) & 79.23 (81.74/76.88) \\
            Top-Neg$^\Diamond$  & 80.55 (81.07/80.23) & 52.86 (52.30/53.55) & - & - & - & 80.39 (82.09/78.90) \\
            CLIM$^\Diamond$ &      85.4 (-/-) & 53.8 (-/-) & 69.6 (-/-) & 70 (-/-) & 67.9 (-/-) & -\\
            \hline
            \multicolumn{6}{l}{\textbf{DS-NER without Self-training}}\\
\hline
            KB-Matching$^\#$    & 71.40 (81.13/63.75)   & 35.83 (40.34/32.22)   & 59.51 (63.86/55.71)   & 47.76 (47.90/47.63)  & 52.45 (62.59/45.14) & 64.32 (\textbf{86.39}/51.24)$^\dagger$ \\
            AutoNER$^\#$           & 67.00 (75.21/60.40)   & 26.10 (43.26/18.69)   & 67.18 (64.63/\underline{69.95})   & 47.54 (43.54/52.35) & 51.39 (48.82/54.23)  & \underline{79.99} (\underline{82.63}/77.52)$^\dagger$ \\
            RoBERTa-ES$^\#$    & 75.61 (\underline{83.76}/68.90)    &   46.61 (\underline{53.11}/41.52)  & \textbf{68.11} (\textbf{66.71}/69.56)    & 51.55 (49.17/54.50)  & 59.11 (60.14/58.11) & 73.66 (80.43/67.94)$^\dagger$ \\ 
            Conf-MPU$^\dagger$            & 79.16 (78.58/79.75) & - & - & - & - & 77.22 (69.79/\textbf{86.42})$^\dagger$ \\
            \textbf{CuPUL-curr}  &      \underline{83.18} (83.69/\underline{82.68}) & \underline{50.12} (47.48/\underline{53.07})  & 67.76 (65.66/\textbf{70.00}) & \underline{66.43} (\underline{58.89}/\textbf{76.18}) & \underline{65.15} (\underline{62.89}/\underline{67.57}) & 79.91 (75.07/85.43 ) \\ 
		  \textbf{CuPUL}           & \textbf{85.09} (\textbf{84.64}/\textbf{85.53})    & \textbf{54.34} (\textbf{54.47}/\textbf{54.20})  &  \underline{68.06} (\underline{66.31}/69.91)  & \textbf{70.53} (\textbf{67.06}/\underline{74.39})      & \textbf{73.10} (\textbf{74.65}/\textbf{71.62})  & \textbf{80.19} (74.91/\underline{86.28}) \\
        
          \hline
          \multicolumn{6}{l}{\textbf{DS-NER with Self-training}}\\\hline
            BOND$^\#$            & 81.15 (82.00/80.92)     & 48.01 (53.16/43.76)   & 68.35 (\underline{67.14}/69.61)   & 60.07 (53.44/68.58) & 65.74 (67.37/64.19) & - \\
            RoSTER$^\P$          & 85.40 (85.90/84.90)      & -                     & -            & \underline{67.80} (\underline{64.90}/\underline{71.00})      & -          & - \\
            SCDL$^\ddag$            & 83.69 (\textbf{87.96}/79.82)   & 51.10 (59.87/44.57)    & 68.61 (\textbf{67.49}/69.77)   & 64.13 (62.25/66.12)  & 68.47 (68.71/68.24)  & - \\
            ATSEN$^\ddag$ &   85.59 (86.14/85.05)   & \underline{52.46} (\textbf{62.32}/45.30)   & \underline{68.95} (66.97/\underline{71.05})   & -  & 70.55 (71.08/70.03)  & - \\
            desERT$^\ddag$ &  \textbf{86.95} (\underline{86.41}/\textbf{87.49})   & 52.26 (\underline{57.65}/\underline{47.80})   & \textbf{69.17} (66.63/\textbf{71.92}) & 65.99 (62.87/69.42)  & \underline{72.73} (\underline{72.48}/\underline{72.97})  & - \\
            
            \textbf{CuPUL+ST} & \underline{86.64} (86.02/\underline{87.27})  &  \textbf{54.78} (57.32/\textbf{52.46})  & 68.20 (66.57/69.11)     & \textbf{70.19} (\textbf{66.96}/\textbf{73.7}4)  & \textbf{74.48} (\textbf{76.06}/\textbf{72.97})  & \textbf{80.92} (\textbf{75.45}/\textbf{87.26}) \\
		\hline
        \end{tabular}
        }
        \caption{Performance on benchmark datasets: F1 Score (Precision/Recall) (in \%). $\textbf{\#}$ marks the row of results reported by \citet{liang2020bond}. $\textbf{\P}$ marks the row of results reported by \citet{meng2021distantly}, where results for Twitter, OntoNote5.0 and Webpage are not reported in \citet{meng2021distantly}. $\textbf{\ddag}$ marks the row of results reported by \citet{zhang2021improving}. $\Diamond$ marks the row of results from the method proposed paper respectively. $\dagger$ marks the results from \citet{zhou2022distantly}. The best results are in \textbf{bold}, second best results are in \underline{underline}.}\label{tab:table_benchmark}
\end{table*}

\section{Ablation Study} \label{append:ablation}


\textbf{Curriculum Learning.} To evaluate the effectiveness of curriculum learning in \our, we compare it with two variations of itself. First, we use the five voters trained using positive and sampled negative examples and take the average of their soft label predictions as the result. The results are shown as voter ensemble in Table \ref{tab:table_ablation}. 
Second, we include the result of CuPUL-curr from Table \ref{tab:table_benchmark} since it is another variation. 
To evaluate the effectiveness of the Conf-MPU loss estimation for curriculum learning in \our, we use the regular loss estimation, which considers unlabeled tokens as non-entity tokens, denoted as w/o Conf-MPU in Table \ref{tab:table_ablation}.

Our analysis reveals the critical role of each component, as removing any of them results in a significant drop in the F1 score. Compared CuPUL-curr with w/o Conf-MPU, we find that CuPUL-curr consistently achieves higher recall. This is attributed to Conf-MPU primarily addressing false negatives \cite{zhou2022distantly} and partial false positives (see the following discussions), leading to more tokens being predicted as entities, thereby enhancing recall. Conversely, w/o Conf-MPU exhibits higher precision since it tackles both false positives and positive type errors. Addressing positive type errors benefits both precision and recall, but the increase in precision is more pronounced compared to CuPUL-curr.

We observed an interesting synergistic effect on the Wikigold dataset: CuPUL has much higher precision than w/o Conf-MPU and CuPUL-curr, as shown in Table \ref{tab:table_ablation}.
To investigate this phenomenon further, we examined the loss function of Conf-MPU. For clarity, we denote the four terms in Eq.(\ref{eq8}) as follows.

\vspace{-1mm}
\begingroup\makeatletter\def\f@size{9}\check@mathfonts
\def\maketag@@@#1{\hbox{\m@th\large\normalfont#1}}%
\begin{align} \label{eq_abcd}
    A &= \ell (f(x_{j}^{T_i^*}, \boldsymbol{\theta}), i) \nonumber \\
    B &= \mathds{1}_{\hat{\lambda}(x_{j}^{T_i^*}) > \epsilon} \ell (f(x_{j}^{T_i^*}, \boldsymbol{\theta}), 0)\frac{1}{\hat{\lambda}(x_{j}^{T_i^*})} \nonumber \\
    C &= \ell (f(x_{j}^{T_i^*}, \boldsymbol{\theta}), 0) \nonumber \\
    D &= \mathds{1}_{\hat{\lambda}(x_{j}^{T_0^*}) \leq \epsilon} \ell (f(x_{j}^{T_0^*}, \boldsymbol{\theta}), 0) 
\end{align}\endgroup

If a sample is annotated as an entity of a certain type, the Conf-MPU loss on this token is $A+B-C$. If the confidence score for this token is lower than $\epsilon$, then $B=0$ and the Conf-MPU loss on this token is $A-C$. While using regular non-PU-based loss, the loss of this sample is $A$. For a false positive sample, if Conf-MPU also has a low confidence score, and the loss on this sample $A-C$ is smaller than $A$ (the regular loss). Consequently, Conf-MPU can avoid overfitting to false positive errors for such cases. Conf-MPU cannot handle samples with positive type errors. For those samples, Conf-MPU may still have high confidence scores that they are entities (close to 1), leading to $B-C$ close to $0$, and thus the loss is $A$, same with regular loss. So, in summary, Conf-MPU can be robust to false positives (non-entity samples labeled as entities) and false negatives (entity samples mistakenly labeled as non-entity) but not to positive type errors (e.g., a sample of type PER is labeled as ORG). In \cite{zhou2022distantly}, since they assume all positive annotations are correct, only the impact of false negatives was discussed.

Curriculum learning, on the other hand, handles false positives and positive type errors by learning from cleaner samples earlier and with more epochs. We also noticed that the three error types may be of different difficulty scores in our curriculum scheduler. Some false positive entities in Wikigold, such as “The” and “Welcome”, have relatively low difficulty scores because voters agreed that they are not entities. This type of noise was introduced in the 2nd and 3rd curriculum, resulting in a bigger impact than noise introduced in later curricula. When Curriculum learning and Conf-MPU are combined together, the false positive noises introduced in early curricula, which had low $\lambda$, can be successfully addressed by the Conf-MPU loss function. This significantly improves model precision and creates a synergistic effect on the Wikigold dataset. Twitter, on the other hand, is dominated by false negatives (60.41\%). Curriculum learning without Conf-MPU suffered from the false negatives more, resulting in low recall. The Conf-MPU loss in CuPUL addressed this error issue and, therefore, improved recall.

\textbf{Distant Labels.} In previous methods, a moderately well-trained model is often used to detect label noise, and the confidently predicted soft labels from the moderately well-trained model are often used to replace the noisy distant labels. Based on our previous experiments, the ensembled voters can be viewed as a moderately well-trained model, and the earlier curricula are formed with data that the moderately well-trained model can confidently predict. We study which labels should be used for curriculum learning in \our, the voters' ensembled soft labels or the noisy distant labels. Note that the ensembled labels used here are the soft labels of the voters' ensemble. We use KL-divergence as the loss function in curriculum learning to learn from soft labels.

Figure \ref{fig: ensemble vs cu} plots the results regarding F1 scores on test data with respect to incremental curriculum stages. We can see that CuPUL learns in almost all stages of the curricula, and the F1 value is steadily improving until the second last curriculum. However, using ensembled soft labels, the model has a good start but reaches the upper bound quickly. We have the following insights from this experiment. 1) A model that only learns from the confidently predicted labels and ignores the potential noisy data may converge faster but can be impacted by the performance bottleneck of the initial model. 2) the last curricula may contain high label noise, so training on the last curricula may degrade the performance slightly. However, thanks to the curriculum learning schedule, the model is overall robust to noise in the last curricula.

\begin{figure}[h]
  \centering
  \includegraphics[width=0.7\linewidth]{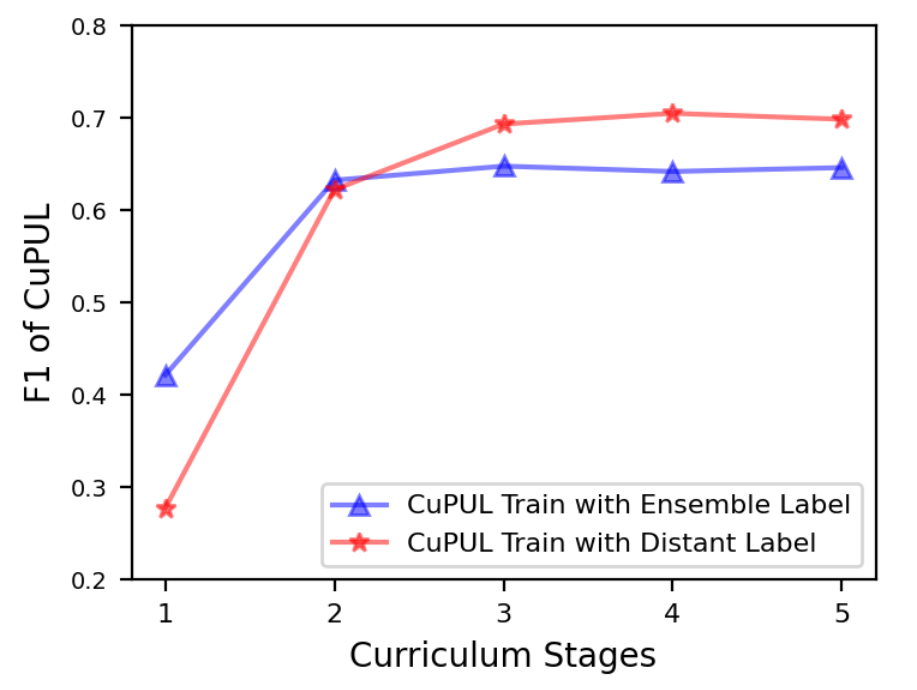}
  \caption{\small{F1 scores of \our on test data of Wikigold trained with Distant Labels (red) and Ensembled Labels from voters (blue) after each curriculum training stage.}}
  \label{fig: ensemble vs cu}
\end{figure}

\section{Parameter Study}\label{section: para_study}

Here, we perform parameter studies. Due to the simplicity of \our, we mainly study two parameters: the number of voters $V$ and the number of curricula $\eta$. To ensure comparability of experimental results, we keep all other parameters fixed and only change the corresponding parameter ($V$ or $\eta$) to demonstrate their impact. The experiments are carried out on Wikigold.

\begin{table*}[htbp]
    \centering
    \resizebox{0.9\textwidth}{!}{
    \begin{tabular}{ccccccccccccccccccccc}
    \hline
            \textbf{Index}     & 1&2&3&4&5&6&7&8&9 \\
            \textbf{Token}       &  the      & regiment  & was   & attached  & to	  &Howe   &'s     & Brigade	& of  & $\cdots$	 \\
         \hline
         \textbf{Ground Truth} &  O       &     O     &  O    &  O        & O        & ORG   & ORG   & ORG   & O  \\
         \textbf{Distant Label} &  O       &     O     &  O    &  O        & O        & ORG   & ORG   & ORG   & ORG  \\
         \textbf{Curriculum \#} &0&0&0&0&0&2&3&2&4 \\
         \hline
         \hline
         \textbf{Index}     & 10&11&12&13&14&15&16&17&18 \\
         \textbf{Token} & the & IV & Corps & of & the &	Army & of & the & Potomac\\
         \hline
         \textbf{Ground Truth} &  O       &     ORG     &  ORG    &  O        & O        & ORG   & ORG   & ORG   & ORG  \\
         \textbf{Distant Label} &  O       &     ORG     &  ORG    &  ORG      & O        & ORG   & ORG   & O   & O \\
         \textbf{Curriculum \#} &0&2&2&2&0&2&2&0&0 \\
         \hline
    \end{tabular}
    }
    \caption{Case study on Wikigold. The selected sentence is "After burying the dead on the field of Second Battle of Bull Run, the regiment was attached to Howe 's Brigade of Couch 's Division of the IV Corps of the Army of the Potomac where it replaced De Trobriand 's 55th New York, Gardes Lafayette regiment on September 11, 1862." This table shows two pieces of this sentence.}\label{tab:case study}
\end{table*}

\subsection{Number of Voters $V$}
\begin{figure}[t]
  \centering
  \includegraphics[width=0.7\linewidth]{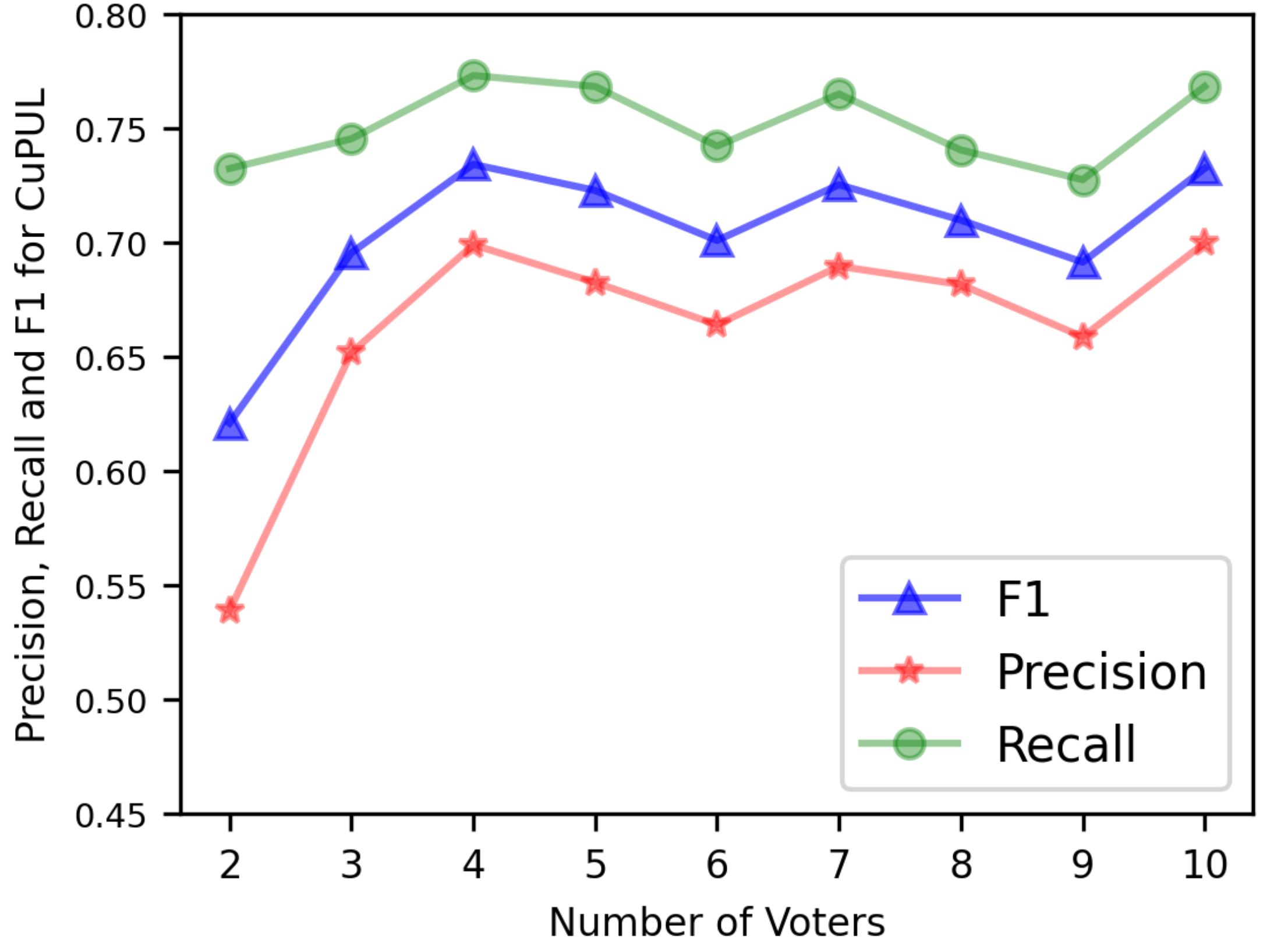}
  \caption{Span Level Precision, Recall, and F1 scores of \our with respect to Number of Voters $V$. }
  \label{fig: voter number}
\end{figure}
Figure \ref{fig: voter number} shows the effect of the number of voters $V$ to \our performance. From the figure, we can see that when there are only two voters, the performance of \our is poor. This is understandable because, with too few voters, the difficulty scores estimated are unreliable, which leads to a low-quality curriculum scheduler. As the number of voters increases, the performance of \our also rapidly improves. When the number of voters is 4, it reaches a local maximum. Then, as the number of voters increases, the new voters can no longer provide new information for difficulty estimation, and the results of \our are stabilized around 0.7. Therefore, with the consideration of computation efficiency, a moderate number greater than or equal to 4 can be chosen for the number of voters.

\subsection{Number of Curricula $\eta$}
\begin{figure}[t]
  \centering
  \includegraphics[width=0.7\linewidth]{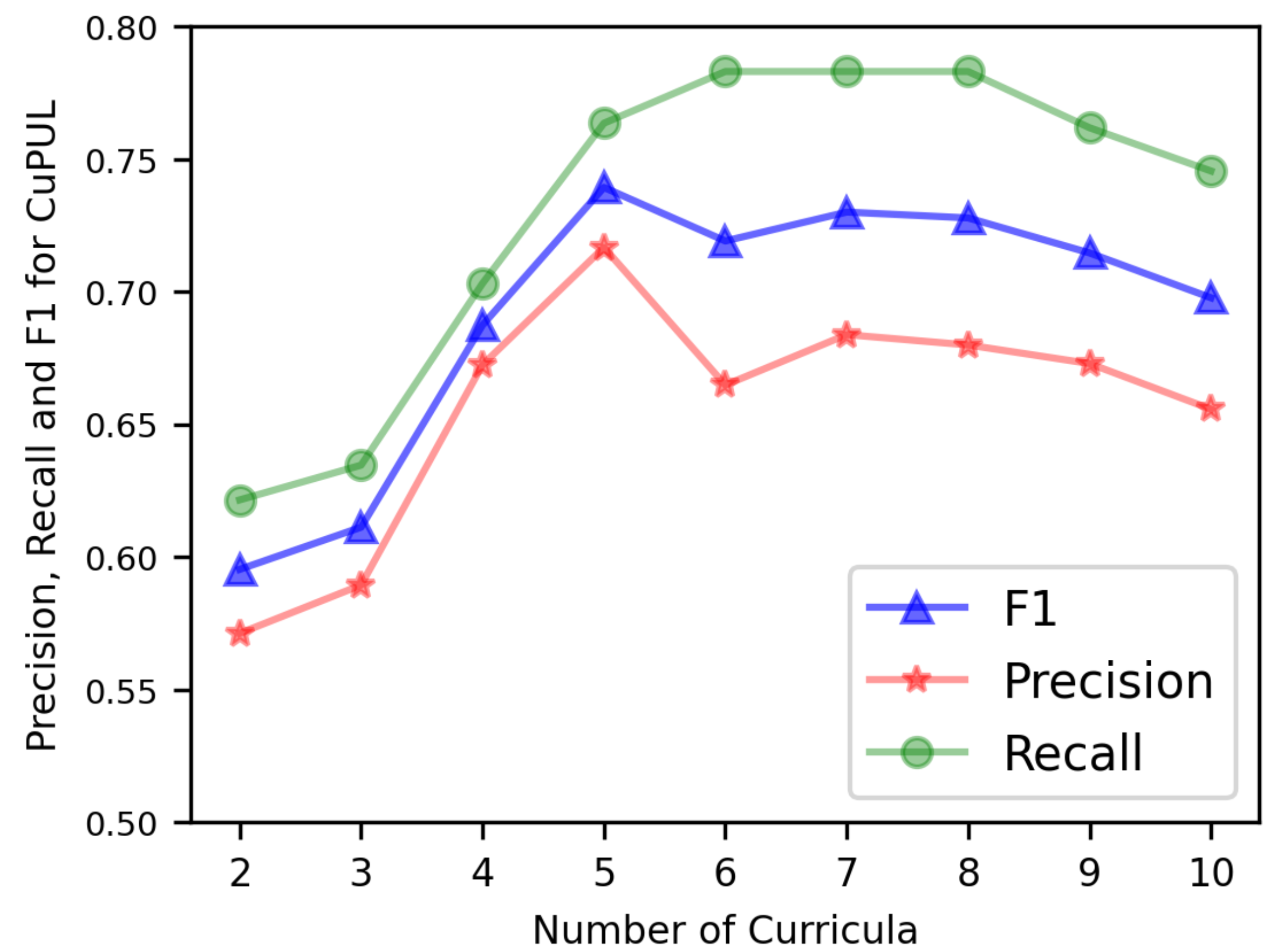}
  \caption{Span Level Precision, Recall, and F1 scores of \our with respect to Number of Curricula $\eta$.}
  \label{fig: curriculum number}
\end{figure}
Figure \ref{fig: curriculum number} shows the effect of the number of curricula to \our performance. Like the number of voters, when the number of curricula is small, the performance of \our is poor. Too few curricula can reduce the ability to distinguish between easy and difficult tokens, leading to ineffective curriculum learning. With the increase of $\eta$, the performance of \our also improves and reaches the best performance at $\eta=5$. After that, as the number of curricula increases, the performance of \our is relatively stable. The performance of \our begins to decline after $\eta>8$. The decline may be caused by the data having been trained too many rounds, and the model starts to overfit to noisy labels.

\section{Efficiency Analysis} \label{appendix:effi}
\begin{table}[t]
    \resizebox{0.5\textwidth}{!}{%
    \begin{tabular}{c|cccccc}
    \hline
         \textbf{} & \textbf{BOND} & \textbf{RoSTER} & \textbf{SCDL} & \textbf{Conf-MPU} & \textbf{CuPUL} & \textbf{CuPUL-ST} \\
         \hline
         \multirow{2}{*}{Run Time}
         & 978s & 2397s & 4319s  & 732s & 819s & 1733s \\
         & 16m18s & 39m57s  & 71m59s & 12m12s & 13m39s & 28m53s \\
         \hline
    \end{tabular}
    }
    \caption{Efficiency analysis on CoNLL03, m means minute, s means second}\label{tab:efficiency}
\end{table}

In order to evaluate the efficiency of \our, we undertook performance timing of the principal methods on CoNLL03, with the results displayed in Table \ref{tab:efficiency}. All tests were performed on an identical computing infrastructure. The training epochs for BOND and SCDL were preset to 5, while the parameter configurations for RoSTER adhered strictly to those detailed in their respective paper. The data in the table reveals that Conf-MPU had the least time requirement. Our approach, \our, demonstrated competitive performance in this regard. Even when the self-training procedure was incorporated into CuPUL-ST, it maintained a substantial efficiency advantage relative to both RoSTER and SCDL.

\section{Case Study} \label{appendix: case_study}

To gain an intuitive understanding of how the curriculum helps \our, we selected a sentence from the Wikigold corpus to show how \our learns. As shown in Table \ref{tab:case study}, we give the tokens, ground truth labels, the distant labels, and the Number of Curricula for each token in the sentence. We assign each token an index for ease of discussion. We display a sentence in two lines and omit some repeated parts. As can be seen from Table \ref{tab:case study}, 
the two ``of'' (token 9 and token 16) are learned in different curricula. The one with the false positive label (token 9) is arranged in the fourth curriculum, whereas the one with the correct label (token 16) is learned early (the second curriculum). This shows that the pre-trained language model has the capability of providing prediction results for each token while retaining context information, and thus, the difficulty scores can be determined at the token level.

\end{document}